# EXPLOITING BERT TO IMPROVE ASPECT-BASED SENTIMENT ANALYSIS PERFORMANCE ON PERSIAN LANGUAGE

Hamoon Jafarian[1], Amirhosein Taghavi[2], Alireza javaheri[3], Reza Rawassizadeh[4]

[1] Independent Researcher, hamoon.jafarian@gmail.com
[2] Independent Researcher, amir.h.t.1987@gmail.com
[3] Independent Researcher, alirjavaheri@gmail.com
[4] Boston University, Boston, MA, USA, rezar@bu.edu

## ABSTRACT

*Aspect-based sentiment analysis (ABSA) is a more detailed task in sentiment analysis, by identifying opinion polarity toward a certain aspect in a text. This method is attracting more attention from the community, due to the fact that it provides more thorough and useful information. However, there are few language-specific researches on Persian language. The present research aims to improve the ABSA on the Persian Pars-ABSA dataset. This research shows the potential of using pre-trained BERT model and taking advantage of using sentence-pair input on an ABSA task. The results indicate that employing Pars-BERT pre-trained model along with natural language inference auxiliary sentence (NLI-M) could boost the ABSA task accuracy up to 91% which is 5.5% (absolute) higher than state-of-the-art studies on Pars-ABSA dataset.*

## 1. INTRODUCTION

Sentiment analysis or opinion mining is the computational study of people's opinions, sentiments, emotions, appraisals, and attitudes towards entities such as products, services, organizations, individuals, issues, events, topics, and their attributes (Pei, Sun et al. 2019). Nowadays, there are many applications for sentiment analysis. Industry use sentiment analysis of online reviews about their product to find which aspects to improve. As well as, finding the real needs of market and smart marketing. Therefore, having a sense of general sentiment about a service or product could be used as a valuable guide for businesses to move in right direction (Ataei, Darvishi et al. 2019). Another application of the sentiment analysis (SA) is in politics where political entities could define their campaign strategy based on public opinion. Sentiment analysis on social media is a powerful tool for determining public opinion (Ramteke, Shah et al. 2016). Moreover, stock market prediction based on public sentiments expressed on social media has been an interesting application of the SA (Pagolu, Reddy et al. 2016). These and many other emerging applications of the SA, such as market potential analysis (Javaheri, Moghadamnejad et al. 2020) makes it an intriguing and crucial field of research.

Sentiment analysis can be broadly classified into three categories: document-level sentiment analysis, sentence-level sentiment analysis and aspect-based sentiment analysis (Pei, Sun et al. 2019). Document-level and sentence-level sentiment analyses allocate a sentiment polarity toward the whole document or sentence. In many cases, this could be useful, but the superiority of aspect-based sentiment analysis (ABSA) becomes vivid when we have different aspects mentioned in sentence or a document. For instance, consider a review about a mobile phone such as "The screen is awesome but the battery life is too short", document-level or sentence-level sentiment analysis fail to provide a useful result while aspect-based sentiment analysis can be applied here to identify the aspects and sentiments towards the identified aspects. Sentiment analysis was acknowledged in early 2000 (Turney 2002) and research in this area kept growing for different applications and methods ever after.

Among different employed methods for sentiment analysis, deep neural networks showed promising results (Do, Prasad et al. 2019) and the advent of Bidirectional Encoder Representations from Transformers (BERT) lead to the state-of-the-art outcome (Devlin, Chang et al. 2018), that is available to the public. Although Generative Pre-trained Transformer 3 (GPT-3) which is a neural-network-based language model, has also shown promising results, at the time of writing this paper, it is not available for the public access (Brown, Mann et al. 2020).

However, most research works in this area focus on English language. Thus, developing ABSA for non-English languages is not widely explored. The aim of this research is to present an aspect-based sentiment analysis for Persian language using pre-trained BERT model with sentence pair input to improve the previous researches in this field.

## 2. RELATED WORKS

(Sun, Huang et al. 2019) presented that since the form of the input in BERT model could be a single text (sentence) or a pair of texts (sentences), ABSA task can be converted into a sentence-pair classification task which enhances ABSA task considerably. In this method, the first sentence is the text containing sentiment toward one or more aspects and the second sentence (auxiliary sentence) includes the specific aspect that we are interested in. Pre-trained BERT model takes this sentence-pair as the input and performs a sentence-pair classification. ABSA using BERT was investigated further (Hoang, Bihorac et al. 2019, Li, Bing et al. 2019) and results showed that it could be considered one of the best available tools for the ABSA task.

However, the mentioned studies were solely for English language. Multilingual BERT model was introduced for some languages other than English but it usually does not present the same quality mainly because it is not precisely trained over language-specific vocabulary. (Farahani, Gharachorloo et al. 2020) proposed a monolingual Persian language model (ParsBERT) based on BERT architecture and trained and fine-tuned on a wider Persian language source. To test and compare the performance of ParsBERT model, different datasets including SentiPers sentence level datasets were used. It was concluded that ParsBERT model could improve sentiment analysis and even outperform multilingual BERT model.

There are a few datasets available in Persian for document and sentence level sentiment analysis (Hajmohammadi and Ibrahim 2013, Bagheri and Saraee 2014, Golpar-Rabooki, Zarghamifar et al. 2015). However, the first aspect-level dataset for Persian language was presented by (Hosseini, Ramaki et al. 2018), SentiPers, containing more than 26,000 sentences with 21,375 aspects.

(Dastgheib, Koleini et al. 2020) used SentiPers dataset on sentence level for sentiment text classification. They proposed a hybrid method of structural correspondence learning (SCL) and convolutional neural network (CNN). The results showed the proposed method works better than a simple CNN and classic Naïve-Bayes classifier. (Sharami, Sarabestani et al. 2020) also used sentence-level SentiPers dataset for sentiment analysis. Bidirectional LSTM and CNN were the two investigated deep learning architectures in the research. Their results showed both methods work better than baseline models and B-LSTM has the best results.

(Ataei, Darvishi et al. 2019) presented a new aspect-based manually annotated Persian dataset, Pars-ABSA, containing 5602 unique reviews including 10002 number of aspects (targets). In addition, the paper presented a baseline for ABSA using state-of-the-art methods with a focus on deep learning. To the best of our knowledge, (Ataei, Darvishi et al. 2019) is the only available research on ABSA for Persian language. Therefore, in the present paper, the Pars-ABSA dataset and its presented baseline are used to compare the proposed method performance.

The novelty of the present paper is using pre-trained BERT with sentence pair input for an aspect-based sentiment analysis task on a Persian language database for the first time. Fine-tuning pre-trained BERT model, new state-of-the–art results were achieved on Pars-ABSA dataset. In addition, the effects of utilizing monolingual Persian language model (ParsBERT) and different types of auxiliary sentence on the model performance are investigated.

## 3. MODEL DEVELOPMENT

Our objective in this research is to detect sentiment expressed towards a given aspect term, in a specified sentence. For each input pair (sentence, aspect term), the output is a single sentiment label: positive, negative or neutral. This is similar to SemEval-2014 task 4 subtask 2 (Pontiki, Galanis et al. 2016).

To address this multi-class classification problem (positive, neutral, negative), we use multilingual pre-trained BERT which includes the Persian (Farsi) language. BERT's model architecture is a multi-layer bidirectional transformer encoder based on the original implementation described in (Vaswani, Shazeer et al. 2017) and released in the tensor2tensor library (Devlin, Chang et al. 2018). The BERT-Base Multilingual Cased[1] has 12 Transformer blocks, 12 self-attention heads, 110M parameters and the hidden layer size is 768 (Farahani, Gharachorloo et al. 2020). Figure 1 shows the overall structure of the proposed model. The BERT embedding layer takes the sentence as input and calculates the token-level representations using the information from the entire sentence. Since we are going to use BERT as text classifier a single linear layer is added on top of it. As input data is fed, the entire pre-trained BERT model and the additional untrained classification layer is trained on our specific task which is called fine-tuning. The probability of each class is calculated using Softmax function and since the problem is a multiclass classification, cross-entropy loss function is employed.

Pars-BERT[2] uses the same architecture as BERT and in order to achieve better performance, tries to improve the used Persian corpus for pre-training the model. (Farahani, Gharachorloo et al. 2020) asserted that their model (Pars-BERT) presents a better performance on Persian texts compared to multilingual BERT model, and in the present paper this claim is investigated for an ABSA task. The developed model is available on a public repository[3].

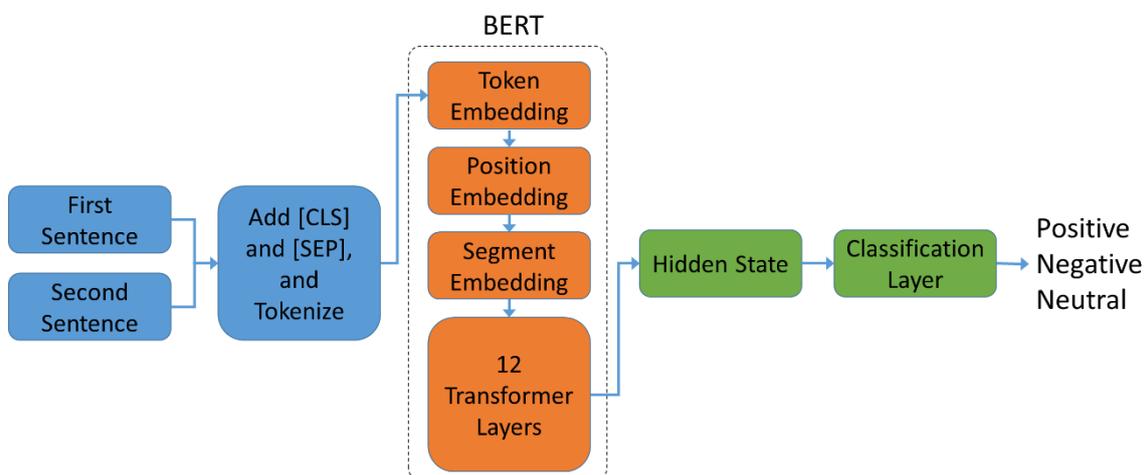

Figure 1 Overall structure of ABSA task using BERT

## 3.1. Dataset

As it has been described, the proposed method is applied on Pars-ABSA dataset[4]. This dataset contains 5,114 targets with positive polarity, 3,061 with negative polarity and 1,827 targets with neutral polarity. Total number of comments is 5,602 which shows that many comments have more than one aspect. Table 1 shows two samples of Pars-ABSA dataset.

Table 1 Samples of Pars-ABSA dataset

| Review | Aspect | Polarity |
|---|---|---|
| این کوله پشتی کیفیت متوسط و تا حدودی قابل قبولی داره ولی به هیچ وجه به این قیمت نمیارزه نهایتا یک سوم این قیمت ارزش واقعیشه (This backpack's quality is medium and acceptable to some extent, but it's not worth the price at all. It's real worth is one third of the price in the best case scenario) | کیفیت (quality) | neutral |
| | قیمت (price) | negative |
| دکمه زومش به جای زوم عکس میگیره ولی رنگش خوبه کیفیتشم خیلی پایینه درکل پیشنهاد نمیکنم (Its zoom button takes pictures instead of zooming, but the color is nice. Its quality is very low, don't recommend it overall) | رنگ (color) | positive |
| | کیفیت (quality) | negative |

## 3.2. Auxiliary Sentence

As it has been described and presented in Figure 1, BERT model receives two sentences (a review and an aspect term) as inputs and predicts the polarity of the review toward the aspect term as output. In other words, the review is fed to the model as the first sentence and the aspect term in form of an auxiliary sentence, would fill the second place. Using the auxiliary sentence (second sentence) enhances the performance of the BERT model for an ABSA task.

There are different arrangements that one could form the auxiliary sentence from aspect term (Sun, Huang et al. 2019). Table 2 shows four different types of auxiliary sentences with examples. In question-answer with multiple output type (QA-M), the auxiliary sentence is a real question and the answer in this case in either positive, negative or neutral. In natural language inference with multiple output (NLI-M), a pseudo-sentence is used as the auxiliary sentence and the output is the same as QA-M type. For binary type auxiliary sentences (QA-B, NLI-B) each target-aspect pair is converted to three sequences and the problem is changed to a binary classification. In NLI-B case, for example, a comment "The food was delicious" is converted to three sequence with the same first sentence and following second sentences: food-positive, food-negative and food-neutral. The output to each is between 0 (no) to 1 (yes). In the mentioned example the model chooses the class with highest score (probability of yes) among the three sequences.

Table 2 Auxiliary sentence types based on (Sun, Huang et al. 2019)

| Method | Auxiliary sentence example | Output example |
|---|---|---|
| QA-M | What do you think of the (aspect) | positive |
| NLI-M | (aspect) | neutral |
| QA-B | The polarity of aspect (aspect) is positive | 0 |
| NLI-B | (aspect)-positive | 1 |

---

[4] https://github.com/Titowak/Pars-ABSA

### 3.3. Hyperparameters

For fine-tuning the models, to reach the highest accuracy we set the batch size to 16, learning rate 2e-5 and number of epochs to 4.

### 4. MODEL EVALUATION

In this section we report the performance in terms of accuracy and F-score in comparison to the state-of-the-art models on Pars-ABSA dataset. (Ataei, Darvishi et al. 2019) presented a baseline for ABSA on Pars-ABSA dataset using 6 recent models with focus on deep learning:

- (Huang, Ou et al. 2018) using an attention-over-attention neural network (AOA)
- (Liu, Zhang et al. 2018) using sentence-level content attention and context attention mechanisms (Cabasc)
- (Tang, Qin et al. 2015) using two target dependent long short-term memory network (TD-LSTM)
- (Ma, Li et al. 2017) using interactive attention networks (IAN)
- (Wang, Huang et al. 2016) using attention-based long short-term memory network (ATAE-LSTM)
- (Chen, Sun et al. 2017) multiple attention mechanism (RAM)

For word embedding a Word2Vec[5] model was trained on all of the comments scraped from Digikala website[6]. Table 3 shows the results of using BERT for aspect-based sentiment analysis on Persian language Pars-ABSA dataset along with the previous researches.

Table 3 Performance evaluation of models for ABSA on Pars-ABSA dataset

| | Model | Accuracy | F1 Score |
|---|---|---|---|
| Previous Researches | Cabasc | 68.1 | 61.9 |
| | ATAE-LSTM | 73.8 | 72.25 |
| | IAN | 76.9 | 75.09 |
| | RAM | 77.35 | 74.96 |
| | AOA | 79.15 | 77.13 |
| | TD-LSTM | 85.54 | 84.4 |
| Present Research | **Pars-BERT-NLI-M** | **91** | **90** |
| | Pars-BERT-QA-M | 90 | 89 |
| | Pars-BERT-NLI-B | 89.4 | 88.2 |
| | Pars-BERT-QA-B | 90.5 | 89.5 |
| | Multilingual-BERT-NLI-M | 87.8 | 86.5 |
| | Multilingual-BERT-QA-M | 87.8 | 86.7 |
| | Multilingual-BERT-NLI-B | 86.8 | 85.6 |
| | Multilingual-BERT-QA-B | 85.5 | 84.5 |

---

[5] https://code.google.com/p/word2vec/
[6] http://www.digikala.com, Based on the terms of Digikala, the information of their website is allowed to be used for non-commercial activities with referring to them.

Among previous models, TD-LSTM (Tang, Qin et al. 2015) presents the best performance in spite of the fact that other models were developed afterward and had shown better performance on English datasets.

The results indicate that utilizing BERT model improves the performance of ABSA in all cases. Moreover, monolingual Pars-BERT model slightly outperforms the Multilingual BERT model. Finally, among different types of auxiliary sentence, NLI-M and QA-M offer the best performance on both Pars-BERT and Multilingual-BERT models, while NLI-B and QA-B auxiliary sentences lead to worst accuracy in Pars-BERT and Multilingual-BERT respectively.

## 5. CONCLUSION

In the present paper, an aspect-based sentiment analysis on Pars-ABSA dataset was performed utilizing sentence-pair input BERT model. The Pars-BERT, monolingual pre-trained BERT and the Multilingual pre-trained BERT models were investigated along with four different types of auxiliary sentences. Pars-BERT pre-trained model in combination with NLI-M auxiliary sentence shows the best performance and increases the accuracy to 91% for an ABSA task on the Pars-ABSA dataset which is more than 5.5% (absolute) higher than the previous best model ( i.e.,TD-LSTM).

## REFERENCES

[1] Ataei, T. S., et al. (2019). "Pars-ABSA: an Aspect-based Sentiment Analysis dataset for Persian." arXiv preprint arXiv:1908.01815.

[2] Bagheri, A. and M. Saraee (2014). "Persian sentiment analyzer: A framework based on a novel feature selection method." arXiv preprint arXiv:1412.8079.

[3] Brown, T. B., et al. (2020). "Language models are few-shot learners." arXiv preprint arXiv:2005.14165.

[4] Chen, P., et al. (2017). "Recurrent attention network on memory for aspect sentiment analysis." Proceedings of the 2017 conference on empirical methods in natural language processing.

[5] Dastgheib, M. B., et al. (2020). "The application of Deep Learning in Persian Documents Sentiment Analysis." International Journal of Information Science and Management (IJISM) 18(1): 1-15.

[6] Devlin, J., et al. (2018). "Bert: Pre-training of deep bidirectional transformers for language understanding." arXiv preprint arXiv:1810.04805.

[7] Do, H. H., et al. (2019). "Deep learning for aspect-based sentiment analysis: a comparative review." Expert Systems with Applications 118: 272-299.

[8] Farahani, M., et al. (2020). "ParsBERT: Transformer-based Model for Persian Language Understanding." arXiv preprint arXiv:2005.12515.

[9] Golpar-Rabooki, E., et al. (2015). "Feature extraction in opinion mining through Persian reviews." Journal of AI and Data Mining 3(2): 169-179.

[10] Hajmohammadi, M. S. and R. Ibrahim (2013). "A SVM-based method for sentiment analysis in Persian language." International Conference on Graphic and Image Processing (ICGIP 2012), International Society for Optics and Photonics.

[11] Hoang, M., et al. (2019). "Aspect-based sentiment analysis using bert." NEAL Proceedings of the 22nd Nordic Conference on Computional Linguistics (NoDaLiDa), September 30-October 2, Turku, Finland, Linköping University Electronic Press.

[12] Hosseini, P., et al. (2018). "SentiPers: A sentiment analysis corpus for Persian." arXiv preprint arXiv:1801.07737.

[13]  Huang, B., et al. (2018). "Aspect level sentiment classification with attention-over-attention neural networks." International Conference on Social Computing, Behavioral-Cultural Modeling and Prediction and Behavior Representation in Modeling and Simulation, Springer.

[14]  Javaheri, A., et al. (2020). "Public vs media opinion on robots and their evolution over recent years." CCF Trans. Pervasive Comp. Interact. 2, 189–205.

[15]  Li, X., et al. (2019). "Exploiting BERT for end-to-end aspect-based sentiment analysis." arXiv preprint arXiv:1910.00883.

[16]  Liu, Q., et al. (2018). Content attention model for aspect based sentiment analysis. Proceedings of the 2018 World Wide Web Conference.

[17]  Ma, D., et al. (2017). "Interactive attention networks for aspect-level sentiment classification." arXiv preprint arXiv:1709.00893.

[18]  Pagolu, V. S., et al. (2016). "Sentiment analysis of Twitter data for predicting stock market movements." 2016 international conference on signal processing, communication, power and embedded system (SCOPES), IEEE.

[19]  Pei, J., et al. (2019). "Targeted Sentiment Analysis: A Data-Driven Categorization." arXiv preprint arXiv:1905.03423.

[20]  Pontiki, M., et al. (2016). "Semeval-2016 task 5: Aspect based sentiment analysis." 10th International Workshop on Semantic Evaluation (SemEval 2016).

[21]  Ramteke, J., et al. (2016). "Election result prediction using Twitter sentiment analysis." 2016 international conference on inventive computation technologies (ICICT), IEEE.

[22]  Sharami, J. P. R., et al. (2020). "DeepSentiPers: Novel Deep Learning Models Trained Over Proposed Augmented Persian Sentiment Corpus." arXiv preprint arXiv:2004.05328.

[23]  Sun, C., et al. (2019). "Utilizing BERT for aspect-based sentiment analysis via constructing auxiliary sentence." arXiv preprint arXiv:1903.09588.

[24]  Tang, D., et al. (2015). "Effective LSTMs for target-dependent sentiment classification." arXiv preprint arXiv:1512.01100.

[25]  Turney, P. D. (2002). "Thumbs up or thumbs down? Semantic orientation applied to unsupervised classification of reviews." arXiv preprint cs/0212032.

[26]  Vaswani, A., et al. (2017). "Attention is all you need." Advances in neural information processing systems.

[27]  Wang, Y., et al. (2016). "Attention-based LSTM for aspect-level sentiment classification." Proceedings of the 2016 conference on empirical methods in natural language processing.